# Reconstruction of the shape of irregular rough particles from their interferometric images using a convolutional neural network

A. Abad[1,*], A. Poux[1], A. Boulet[1], M. Brunel[1]

1: UMR CNRS 6614 CORIA, Avenue de l'Université, BP 12, 76801 Saint-Etienne du Rouvray cedex, France
* Correspondent author: abada@coria.fr

## ABSTRACT

We have developed a convolutional neural network (CNN) to reconstruct the shape of irregular rough particles from their interferometric images. The CNN is based on a UNET architecture with residual block modules. The database has been constructed using the experimental patterns generated by perfectly known pseudo-particles programmed on a Digital Micromirror Device (DMD) and under laser illumination. The CNN has been trained on a basis of 18000 experimental interferometric images using the AUSTRAL super computer (at CRIANN in Normandy). The CNN is tested in the case of centrosymmetric (stick, cross, dendrite) and non-centrosymmetric (like T, Y or L) particles. The size and the 3D orientation of the programmed particles are random. The different shapes are reconstructed by the CNN with good accuracy. Using three angles of view, the 3D reconstruction of particles from three reconstructed faces can be further done.

## 1. Introduction

A wide range of optical techniques enable the characterization of particles in complex flows. The choice depends on the characteristics of the flow and of the particles : size of the particles, particle concentration, velocity of the flow, constraints due to the experimental set-up as the possibility to install optical access. Different interferometric techniques can be particularly efficient as Phase/Doppler Particle Analysis (Bachalo et al. 1984, Albrecht et al. 2013), digital holography to perform 3D reconstructions (Buraga-Lefebvre et al. 2000), or rainbow refractometry to measure the size and the refractive index of droplets (van Beeck et al., 1999), without being exhaustive. Interferometric Particle Imaging has been first developed to measure the size of spherical droplets or bubbles in a flow, with applications in sprays, meteorology, combustion, medicine (König et al. 1986, Glover et al. 1995, Kawaguchi et al. 2002, Damaschke et al. 2002, Brunel et al. 2013, Parent et al. 2022, Grandoni et al. 2023). It could be extended to the characterization of irregular rough particles whose interferometric images are speckle patterns (Brunel et al. 2014, Brunel et al. 2015,

Gonzalez Ruiz et al. 2017, Wu et al. 2021). Particle sizing was the first objective, in particular to measure the size of ice particles. But the reconstruction of the shape of the particles from their interferometric images is now an important subject of research. The Fourier transform of the interferometric patterns is indeed directly linked to the spatial autocorrelation of the shape of the particles (Brunel et al. 2015). Phase-retrieval algorithms offer a solution to perform shape's reconstructions (Shen et al. 2018, Delestre et al. 2021, Fienup, 1982). Nevertheless, they suffer the twin image problem in the case of non-centrosymmetric particles, which tends to make reconstructions more difficult and less accurate in this case (Delestre et al. 2021). We investigate in this study the possibility to use a convolutional neural network to perform particle's shape reconstructions from the interferometric images of either centrosymmetric or non-centrosymmetric particles (Piedra et al. 2019, Zhang et al. 2021, Fan et al. 2022, Abad et al. 2023). Section 2 will present the CNN that has been developed and the construction of the database necessary to train the network. Section 3 will present some reconstructions done with the trained CNN while section 4 will present an exemple of 3D reconstruction combining three reconstructions from three angles of view of the same particle.

## 2. Convolutional neural network developed and construction of a database

The neural network developed is a Convolutional Neural Network (CNN) based on a UNET architecture with residual block modules (constituted of normalization, nonlinear function and convolution) (Zhang et al. 2021). The neural network is thus divided in three parts: the encoder encodes informations decreasing the size of the image and increasing its depth (from 256x256x1 to 32x32x196); the decoder ensures the reconstruction of the particle's shape by decreasing the depth of the image and increasing its size (from 16x16x320 to 256x256x1); the bottleneck links the encoder and decoder parts increasing the depth through a convolution and a residual block (from 32x32x196 to 16x16x320). In addition, there are skip connections between the encoder and decoder parts to ensure the stability of the reconstruction. This network has been implemented using Pytorch 2.0.0. Training has been done on the Austral supercomputer at CRIANN (Normandy, France).

One of the main difficulties is to create a sufficient database to train and test the network. In practice, the CNN will require thousands or tens of thousands of interferometric images. In addition, it is necessary to know the exact shape of each particle whose interferometric image is recorded. Experimentally, it means that the interferometric out-of-focus image and the in-focus image of each particle must be recorded simultaneously. It represents an amount of data that is quite impossible to obtain in the case of a real flow with real particles. The solution that we have chosen is to program pseudo- rough particles, whose shapes are perfectly known, on a Digital Micromirror Device (DMD). We record then the interferometric images that are generated under laser illumination of the DMD. This set-up has been proposed in reference Fromager et al. (2017) and is now currently used to test image processing tools in IPI. There is actually no theoretical model to describe light scattering by irregular rough particles of any shape. But assuming that the particle is covered randomly by asperities that act as point emitters, it is possible to make accurate predictions. The DMD set-up is based on this idea as a "programmed particle" is an ensemble of on-state micromirrors that play the role of scattering asperities, randomly distributed in a global contour. With a completely automated set-up, the acquisition of 18000 interferometric images requires approximately 100 hours without interruption. In the same time, the 18000 shapes programmed on the DMD are stored, each of them being linked to its corresponding interferometric image. With this set-up, the interferometric images are real experimental images with their own noise. In addition, they can be numerically noised to increase the diversity of the database for better learning.

Technically, the database consists of six families of particle's shapes. The particles programmed can either be sticks, crosses, dendrites, or T-, L- and Y-shaped particles. Three of them are centrosymmetric while three of them are not. The size of the particles is random in the range [370microns-1.5mm]. The 3D-orientation of each particle is random too. To ensure the most diverse and realistic database in terms of sizes and orientations, there is no correlation between the size and the orientation of a particle. The database is actually constituted of 18000 pairs of images: 18000 programmed particles (filled), and the corresponding 18000 interferometric images. Figure 1 shows six examples of particles programmed on the DMD. Each sub-figure shows the ensemble of micromirrors programmed on-state, covering a stick (a), a cross (b), a dendrite (c), a L (d), a T (e) or a Y (f). For each of this particle, the corresponding interferometric image is recorded on the CCD sensor and stored in the database. Figure 2 ((a) to (f)) shows the interferometric images corresponding to the programmed particles of figure 1 ((a) to (f) respectively). The CCD sensor is a Thorlabs BC106N-VIS/M camera with dimensions of 1545x1164 pixels, and a pixel size of 6.45 μm. The camera is equipped with a 80mm focus length Nikon objective mounted on an extension ring to obtain a defocused image. The parameters of the set-up are those of reference Delestre et al. 2021.

Once the database is complete, 90% of the images are used for training while 10% are dedicated to the phase test. Training time of the CNN on AUSTRAL supercomputer is approximately 45 hours. We use the Mean Square Error (MSE) loss to quantify the error made between predictions and reference images. Figure 3 shows the evolution of the MSE loss during training as a function of the number of epochs. The light blue and orange curves are the raw data. The dark blue and orange curves are the raw data convoluted with a door function of seven epochs. The blue curves show the error obtained on the training data. The orange curves show the error obtained on the test data. The black curve represents the evolution of the learning rate versus the number of epochs. We can observe the decrease of the loss during the training. Training data and test data present similar evolutions indicating a good generalization of the reconstructions.

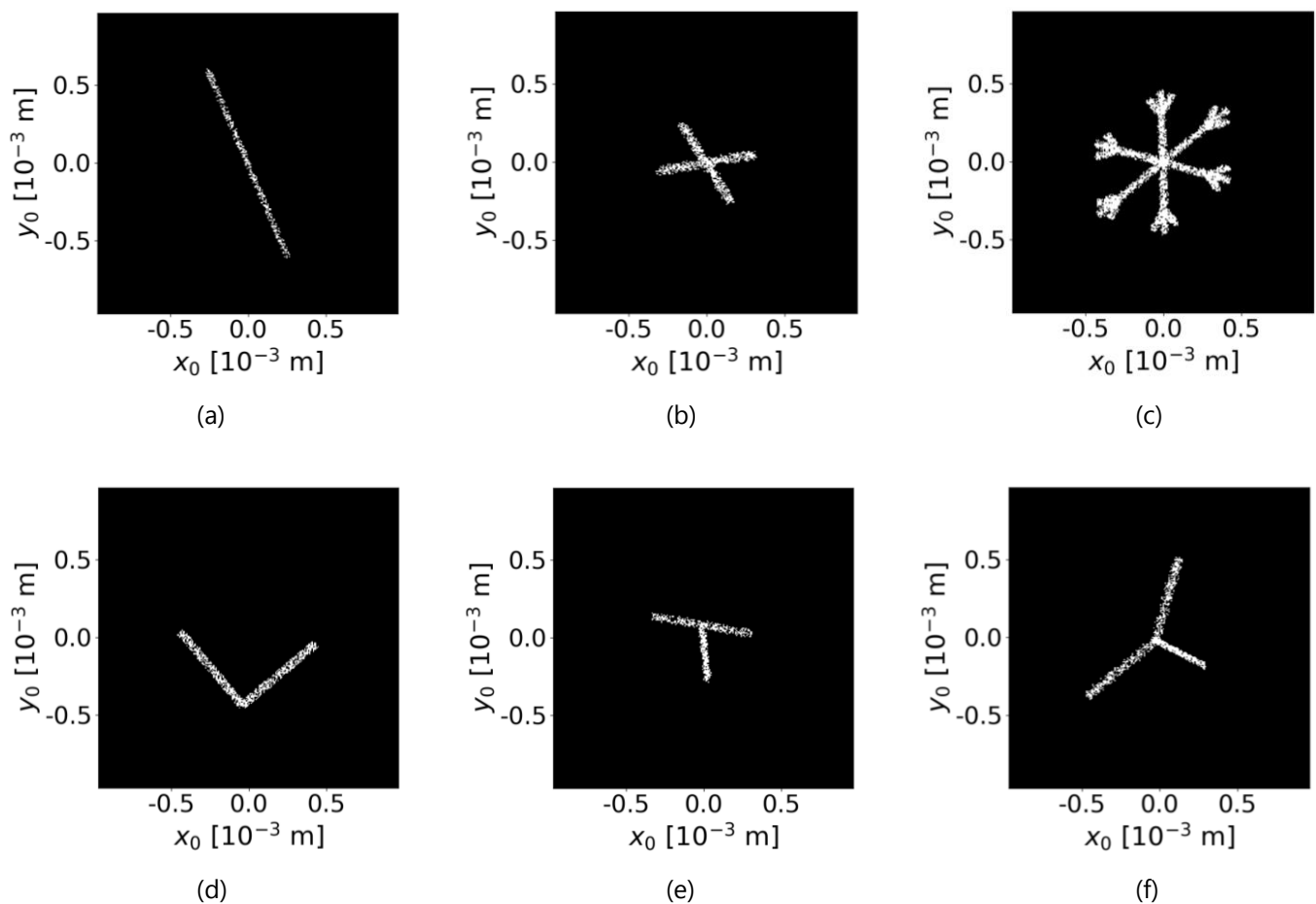

Figure. 1 Examples of the six shapes programmed on the DMD : stick, cross, dendrite, L, T and Y with random sizes and orientations.

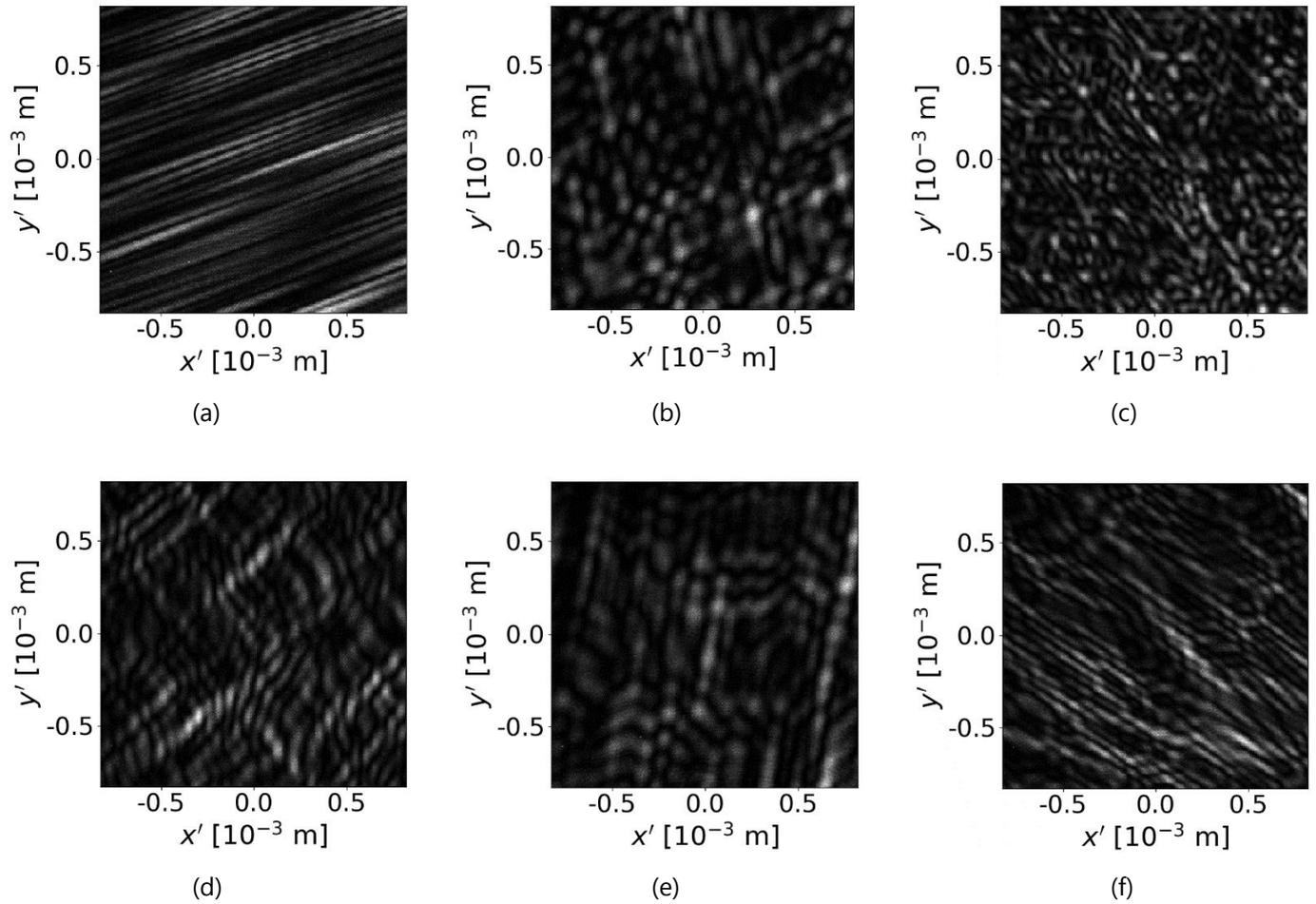

Figure. 2 Interferometric images recorded on the CCD sensors associated to the three particles of figure 1.

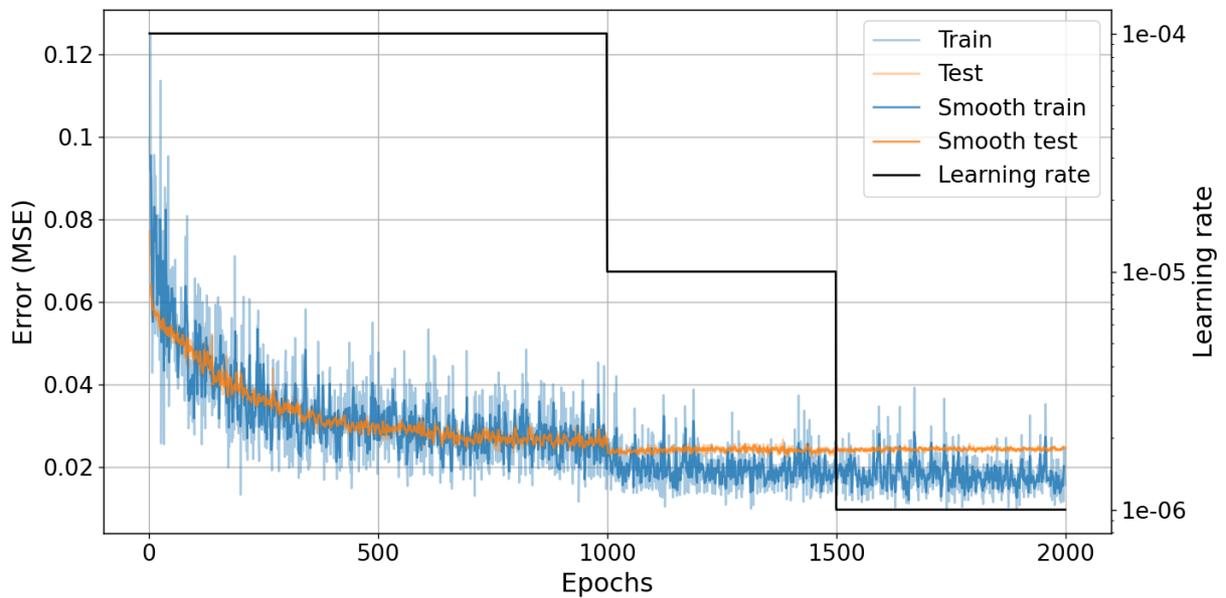

Figure. 3 Evolution of the loss versus the number of epochs.

## 3. Results: shape's reconstructions using the CNN

Figure 4 ((a) to (f)) shows the reconstructions obtained in the case of the six particles of figure 1 ((a) to (f) respectively). Their interferometric images are reported in Fig 2. These six particles belong to the part of the database dedicated to the test by the CNN. Once the training is finished, the reconstruction of the particle's shape from its interferometric image requires only 0.5 second. To compare the reconstructed faces with the original programmed faces, figure 5 ((a) to (f) respectively for all particles analyzed here) shows the difference between the filled programmed shapes and the reconstructed faces. All shapes (programmed and reconstructed) have been normalized to 1. In gray, the difference equals zero. In black or white, it equals -1 or 1. We can observe that the quality of the reconstructions is very good. Note that these reconstructions are neither the best nor the worse. Reconstructions done with a phase-retrieval algorithm in previous works (i.e. the error-reduction algorithm) suffered the twin-image problem in the case of non-centrosymmetric particles. It is not the case with the CNN.

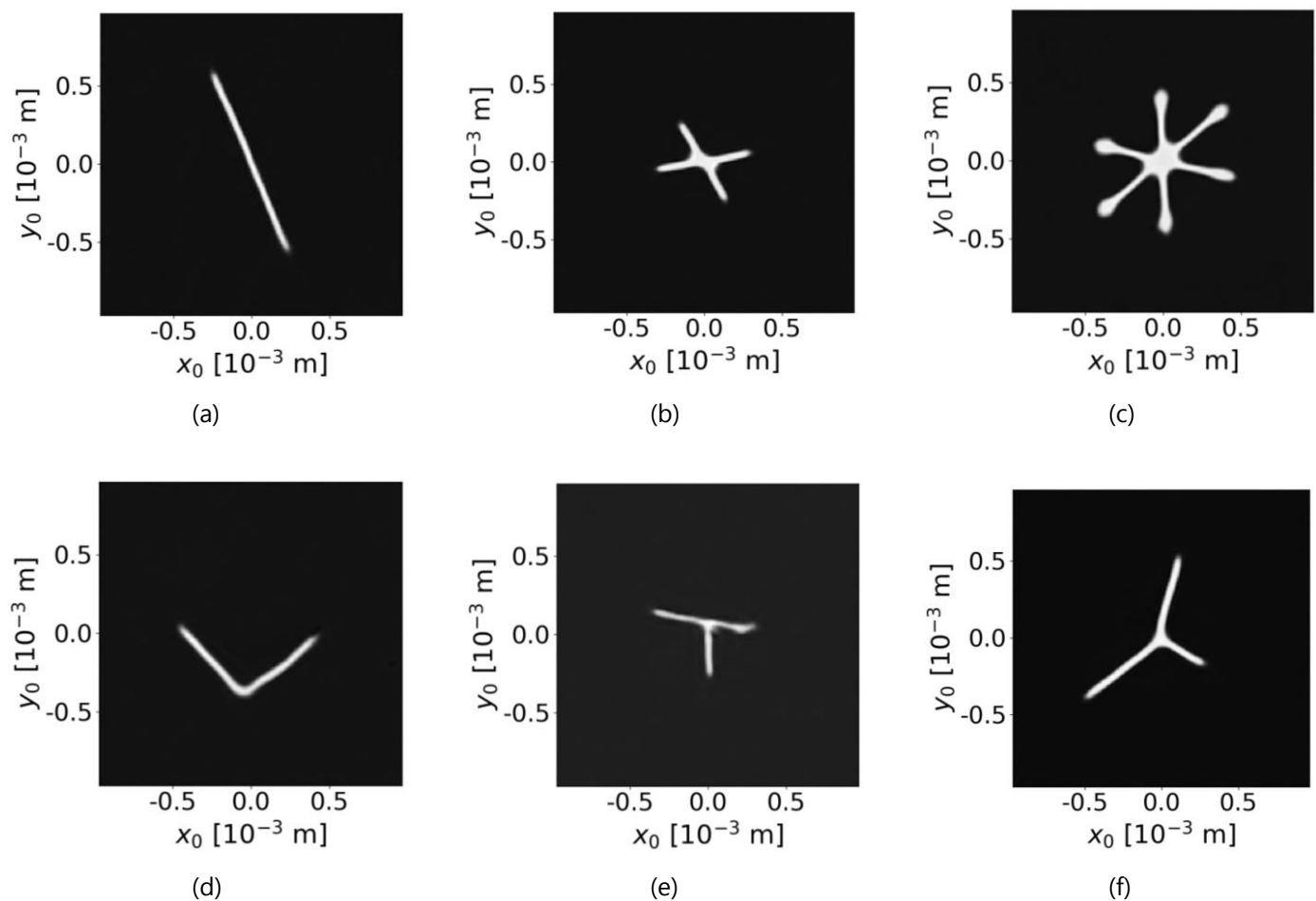

**Figure. 4** Reconstructions of particle's shapes for the three particles of figure 1.

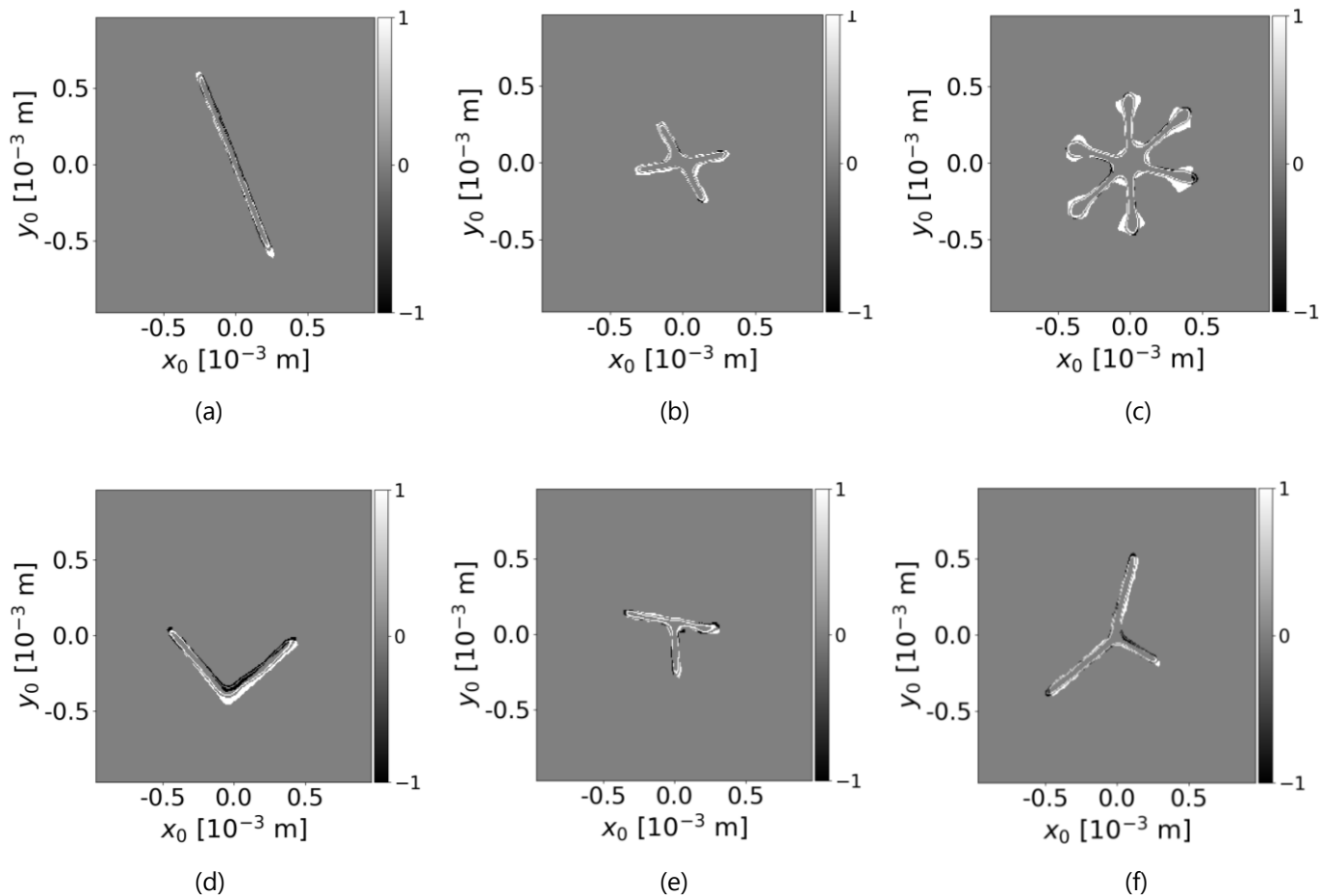

**Figure. 5** Difference between the filled programmed shapes and the reconstructed shapes (for the six particles of figure 1 whose reconstructions have been presented in figure 4).

### 4. 3D reconstruction using multiple views

Using different angles of view, it appears possible to perform the tomography of particles in a flow combining the views (Delestre, Talbi et al. 2021). To illustrate this, figure 6 shows the configuration that will be considered. Three imaging systems are oriented to observe the projections of particles from 3 perpendicular planes: (X,Y), (Y,Z) and (Z,X) respectively. We show here some results obtained with a non-centrosymmetric particle: a T-like rough particle. This case is more difficult using phase-retrieval algorithms, each plane reconstruction suffering the twin-image problem. The filled T-like particle, whose faces will be programmed on the DMD, is reported on figure 7(a) (left part of the figure). The three faces of the particle are programmed successively on the DMD (random choice of active micromirrors located within the contour of figure 7(a)). Their interferometric images have been recorded with the CCD sensor. Figure 8 shows the three reconstructed faces obtained using the CNN. For two of them, the T-shape is visible. For the third one, the orientation of the particle is such that it is observed from the edge. Combining these three views, the 3D-shape obtained after 3D-recombination of the three perpendicular reconstructed views is presented in figure 7(b) (right part of the figure). It shows actually a very good fidelity with the original programmed particle.

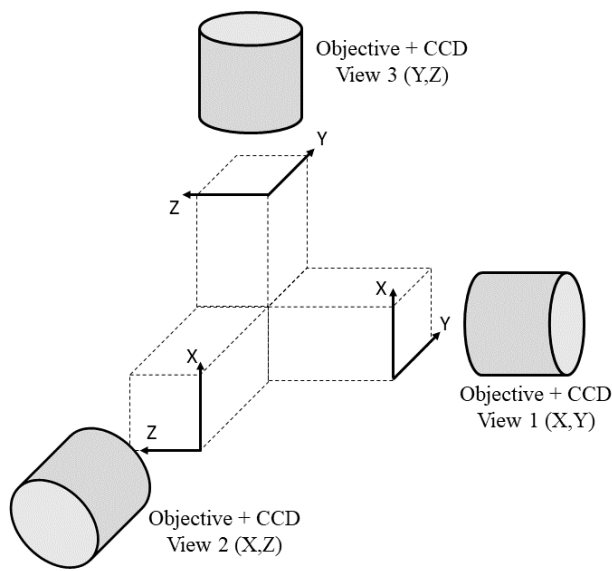

**Figure. 6** Multi-views set-up

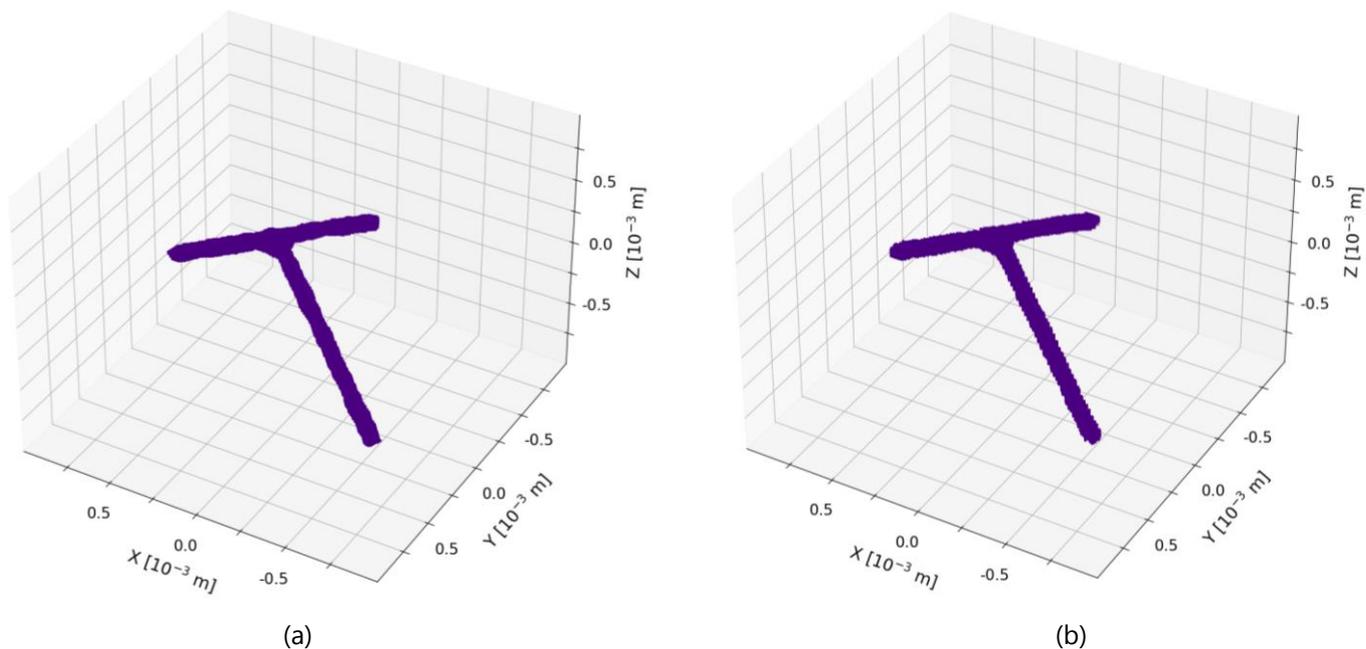

**Figure. 7** Filled T-like particle programmed (a) (left) and 3D reconstruction from recombination of 3 planar reconstructions performed using the CNN (b) (right).

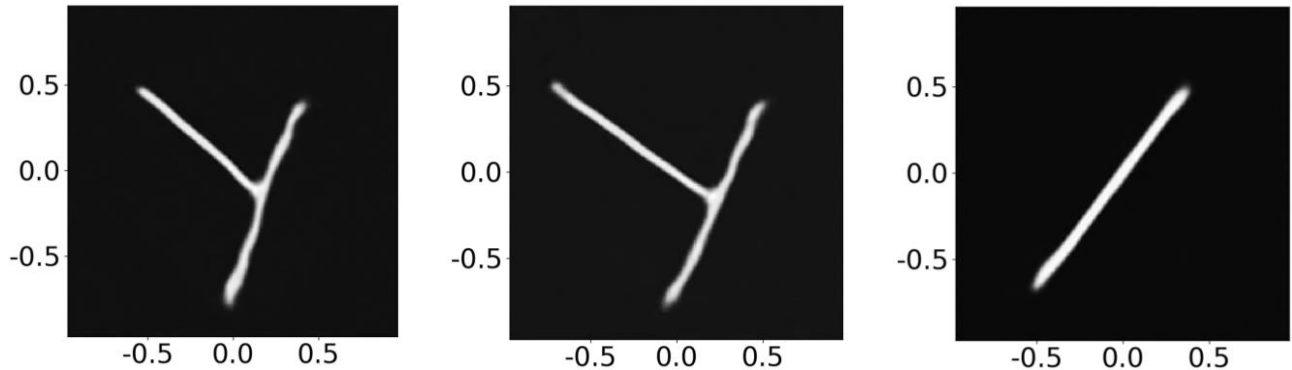

**Figure. 8** Reconstructions of the three views of a T-like particle in the (X,Y), (Z,Y) and (X,Z) planes respectively.

5. Conclusion

A convolutional neural network (CNN) has been developed to reconstruct the shape of irregular rough particles from their interferometric images. A Digital Micromirror Device (DMD) set-up enables the construction of an important bank of experimental interferometric images corresponding to rough objects whose shapes are perfectly known. With this set-up, Pseudo-rough particles are programmed on the DMD that is under laser illumination. The interferometric images are recorded on a CCD sensor in a defocused position. The CNN has shown its ability to reconstruct the shape of either centrosymmetric or non-centrosymmetric particles. Once the learning phase is finished, the reconstruction of a particle's shape from an interferometric image requires 0.5 second. In the future, the bank of images will be enlarged to describe more and more shapes. The use of a CNN should be a powerful tool to perform particle's tomography in the case of multi-views set-ups.

Acknowledgments

The authors would like to acknowledge the Graduate School of Materials and Energy Sciences (GS-MES) for funding.